\RequirePackage{xcolor}
\documentclass[journal,twoside,web]{ieeecolor}

\usepackage{generic}
\usepackage{cite}
\usepackage{amsmath,amssymb,amsfonts}
\usepackage{algorithm}
\usepackage{algorithmic}
\usepackage{graphicx}
\usepackage{textcomp}

\usepackage{booktabs}
\usepackage{multirow}
\usepackage{array}
\usepackage{makecell}

\usepackage{pifont}
\usepackage{balance}
\usepackage{etoolbox}

\usepackage{hyperref}
\providecommand{\refname}{References}

\definecolor{accessblue}{cmyk}{1,0.3,0,0.2}
\def\BibTeX{{\rm B\kern-.05em{\sc i\kern-.025em b}\kern-.08em
    T\kern-.1667em\lower.7ex\hbox{E}\kern-.125emX}}

\begin{document}
\bstctlcite{BSTcontrol}
\title{Role-Guided MOE for Encoder-Level Pathology Representation Learning in WSI Classification}

\author{Xinyu Ma, Xing Yang, Hongtao Jin, Guoquan Zhang, Shijie Zhang, Yu Zhang, and Xitong Li 
\thanks{This work has been submitted to the IEEE for possible publication. Copyright may be transferred without notice, after which this version may no longer be accessible.}
\thanks{Xinyu Ma, Xing Yang, Shijie Zhang, and Xitong Li are with
the Artificial Intelligence Research Institute,
Shenzhen MSU-BIT University, Shenzhen 518172, China
(e-mail: xma200525@gmail.com; yangxing@smbu.edu.cn;
shijie.z@smbu.edu.cn; xli@smbu.edu.cn).}%
\thanks{Hongtao Jin, Guoquan Zhang, and Yu Zhang are with
Shenzhen People's Hospital, Shenzhen 518020, China
(e-mail: 807697204@qq.com; zhang.guoquan@szhospital.com;
zhangyupku2012@163.com).}
\thanks{Corresponding authors: Shijie Zhang; Yu Zhang; Xitong Li.}
}
\maketitle

\begin{abstract}
Whole slide image (WSI) classification is a fundamental task in computational pathology, where the quality of patch representations directly affects downstream aggregation and determines the discriminability of slide-level predictions. Pathology foundation models have recently been widely adopted as frozen feature extractors for WSI classification; however, their fixed encoders may produce patch representations that are insufficiently adapted to target-specific tissue patterns and discriminative cues. Fine-tuning the encoder can improve target adaptation, but it introduces a practical dilemma between pathology-specific representation capacity and adaptation efficiency, particularly in data-scarce pathology settings. To address these challenges, we propose a pathology role-guided mixture-of-experts feed-forward network (MoE-FFN) framework for efficient encoder-level representation learning. We innovatively design a two-stage training paradigm to establish and adapt pathology-aware expert specialization. In source-domain expert initialization, pathology-specific priors are distilled from a frozen Virchow2 teacher into a lightweight DINOv2-small student, while role prototypes serve as weak pathological anchors to encourage distinct expert functions. MoE-FFN blocks are introduced into selected high-level transformer layers to provide transformation diversity for heterogeneous pathological patterns. In target-domain adaptation, the initialized experts are refined through asymmetric prototype-guided optimization, which enhances task-relevant positive evidence and separates confusable hard negatives. The resulting encoder extracts offline patch representations that can be directly integrated with standard MIL aggregators for slide-level classification. Experiments on the public BRACS dataset and a private PAROTID WSI dataset across five representative backbones demonstrate consistent improvements over the strongest baseline.
\end{abstract}


\section{Introduction}
\label{sec:introduction}
\IEEEPARstart{W}{hole} slide images (WSIs) are a cornerstone of digital pathology, providing gigapixel-scale representations that capture diagnostically relevant morphological and structural information~\cite{campanella2019clinical}. Accurate WSI classification is therefore crucial for assisting pathologists in disease diagnosis, staging, and prognostic assessment. Owing to their extremely large image size, mainstream WSI classification methods typically follow the multiple instance learning (MIL) paradigm, in which a slide is divided into patches, patch representations are extracted by a feature encoder, and a MIL aggregator predicts the slide-level label from the resulting patch set~\cite{ilse2018attention,campanella2019clinical,shao2021transmil}. Under this paradigm, the quality of patch representations critically affects downstream aggregation and determines the discriminability of slide-level predictions~\cite{lu2021dataefficient,shao2021transmil}.

Recent advances in pathology foundation models have substantially improved representation learning for histopathological images~\cite{chen2024uni,vorontsov2024virchow,lu2024conch}. Large-scale pretrained models, such as UNI~\cite{chen2024uni} and the Virchow series~\cite{vorontsov2024virchow}, are trained on extensive collections of histopathology images and have demonstrated strong capability in encoding pathological semantics. Consequently, using these models as frozen feature extractors has become a widely adopted strategy for WSI classification tasks~\cite{chen2024uni,vorontsov2024virchow,lu2021dataefficient}. However, since the encoder remains fixed during downstream training, the extracted patch representations may not be sufficiently adapted to a specific target dataset, particularly its task-relevant tissue patterns and discriminative characteristics~\cite{huang2024free,lu2024pathotune}.
This limitation is particularly pronounced in data-scarce pathology domains that are underrepresented in pretraining data or exhibit substantial distribution shifts~\cite{kang2023benchmarking,vorontsov2024virchow,howard2021impact}.
In such cases, the MIL head primarily optimizes slide-level aggregation over fixed patch representations, while the overall classification performance may remain limited by insufficient encoder-level adaptation~\cite{tang2024feature}.
These observations highlight the necessity for effective encoder adaptation to learn target-aware and pathology-discriminative patch representations for WSI classification.

Current encoder fine-tuning strategies for WSI classification face two main limitations. 
First, fine-tuning pathology foundation models is computationally intensive due to their large parameter scale and patch-intensive WSI training, while limited target-domain data may further increase the risk of overfitting~\cite{campanella2019clinical,chen2024uni,vorontsov2024virchow,lee2025benchmarking}. 
In contrast, lightweight vision transformers such as DINOv2-small~\cite{oquabdinov2} are easier to optimize and adapt, but lack the pathology-specific pretraining of foundation models~\cite{chen2024uni,vorontsov2024virchow}.
This creates a practical dilemma between pathology-specific representation capacity and adaptation efficiency.
Moreover, standard encoders apply shared transformations to all patch tokens, which may be insufficient to capture heterogeneous tissue patterns requiring distinct feature mappings. This motivates model architectures that explicitly promote transformation diversity.

To this end, we propose a pathology role-guided mixture-of-experts (MoE) adaptation framework for efficient encoder-level representation learning in low-resource pathology settings.
During the first stage, we develop a novel source-domain distillation strategy using multi-cancer data to learn general pathology-aware representations. A frozen Virchow2 encoder~\cite{vorontsov2024virchow} serves as the teacher to transfer pathology-specific priors through knowledge distillation~\cite{touvron2021training}, while DINOv2-small~\cite{oquabdinov2} is adopted as a lightweight student backbone. We further replace selected high-level feed-forward network (FFN) layers of the student encoder with MoE-based FFN modules, termed MoE-FFNs.
Once optimized, the MoE-FFN blocks can serve as a transferable module across transformer-based encoders through lightweight dimensional alignment, without updating the entire backbone.
To encourage meaningful expert specialization, we introduce role prototypes as weak pathological anchors, inspired by prototype-based representation learning~\cite{snell2017prototypical}. These prototypes guide experts toward distinct tissue-related feature subspaces while discouraging redundant or pathology-irrelevant transformations~\cite{chi2022representation,dai2024deepseekmoe}.
During the second stage, the initialized experts are refined through asymmetric prototype-guided optimization, which aligns positive patches with relevant role prototypes while separating confusable hard negatives from their nearest prototypes. After two-stage training, the MoE-based encoder extracts patch representations offline, which are then aggregated by standard MIL models for slide-level classification.
The main contributions of this work are summarized as follows:
\begin{enumerate}
    \item We design a plug-and-play MoE-FFN module with role-prototype guidance to enable the encoder to learn specialized transformations for heterogeneous pathological patterns. It can replace selected FFN layers in transformer-based encoders while keeping the remaining backbone frozen, and using lightweight projections for dimensional alignment.
    
    \item We develop a two-stage training strategy in which source-domain training enables the MoE-based encoder to learn pathology-aware representations and establish initial expert specialization, while target-domain fine-tuning adapts these representations to distinguish task-relevant positive patterns from confusable negatives.
    
    \item We validate the effectiveness of the proposed method on both public and private WSI datasets, demonstrating consistent performance gains and interpretable representation improvement across diverse settings.
\end{enumerate}
\section{Related Work}

\textbf{MIL for WSI Classification.} Due to the gigapixel resolution of WSIs, most WSI classification methods follow the MIL paradigm~\cite{campanella2019clinical,ilse2018attention}. Early MIL methods used mean or max pooling, while ABMIL introduced attention weights to identify discriminative instances~\cite{ilse2018attention}. CLAM further used class-specific attention and instance-level clustering~\cite{lu2021dataefficient}, and TransMIL modeled correlations among patches with self-attention~\cite{shao2021transmil}. Although MIL aggregators have been widely developed, their performance still depends heavily on patch-level representations. When frozen encoder features are not well adapted to the target task, improving only the aggregation module may still remain limited by input feature quality~\cite{tang2024feature}.

\textbf{Foundation Models for Patch Representation.} Pathology foundation models have advanced patch representation learning by pretraining on large-scale histopathology images and capturing rich tissue morphology, cellular structures, and pathological semantics~\cite{wang2022ctranspath,chen2024uni,vorontsov2024virchow}. Representative models include H\&E-based encoders such as UNI~\cite{chen2024uni} and Virchow~\cite{vorontsov2024virchow}, vision-language models such as CONCH~\cite{lu2024conch}, and recent patch-level, slide-level, or multimodal models such as CTransPath, GigaPath, and TITAN~\cite{wang2022ctranspath,xu2024gigapath,ding2025titan}. In WSI classification, these models are often used as frozen feature encoders to reduce computational cost and overfitting risk on small datasets~\cite{chen2024uni,vorontsov2024virchow}. However, fixed encoder representations may be insufficiently adapted to target specific tissue composition, discriminative cues, and confusing patterns, especially under fine-grained lesions or distribution shifts~\cite{tang2024feature,liu2026hiadapter}. This motivates lightweight and stable optimization of encoder representations for WSI classification.

\textbf{Encoder Optimization.} A direct way to optimize encoders is full or partial fine-tuning, but this is costly for large pathology foundation models and may overfit on limited WSI data. Parameter-efficient fine-tuning (PEFT) methods provide lightweight alternatives, including adapters~\cite{houlsby2019parameter} and LoRA~\cite{hu2022lora}. While widely studied in general
vision and language tasks, these methods do not explicitly model tissue heterogeneity and diverse local morphology in pathology images.
Pathology-specific methods have also been proposed. R$^2$T re-embeds extracted patch features to improve the use of frozen pathology foundation model features in WSI tasks~\cite{tang2024feature}. HiAdapter introduces stain-invariant and morphology-aware adapters with a pathology-prototypical contrastive loss for generalization to unseen cancers and stains~\cite{liu2026hiadapter}. Different from these fine-tuning, PEFT, and post-encoder refinement methods, this work focuses on token transformation inside the encoder, where expert specialization occurs during patch representation formation.

\textbf{Mixture of Experts in Vision and Pathology.} MoE improves model capacity through conditional computation by routing inputs to different expert modules~\cite{jacobs1991adaptive}. In transformers, MoE is commonly introduced by replacing the FFNs with multiple expert FFNs and sparse routing, as in sparse-gated MoE and V-MoE~\cite{fedus2022switch,riquelme2021scaling}. However, MoE training often faces expert collapse, load imbalance, routing instability, and unclear expert specialization~\cite{fedus2022switch,riquelme2021scaling}. These issues are particularly relevant to pathology images, which contain strong tissue heterogeneity and diverse local morphological patterns~\cite{wu2025pamoe,li2025m4}. 
MoE has also been used in computational pathology, mainly in WSI-level MIL aggregation, multi-task prediction, and multi-modal diagnosis. PAMoE proposes a plug-and-play pathology-aware MoE module for tissue-specific feature modeling in MIL pipelines~\cite{wu2025pamoe}. M4 introduces multi-gate MoE into histopathology multi-task MIL for predicting multiple genetic mutations from a single WSI~\cite{li2025m4}. Gated MoE has been used for multi-modal pathology diagnosis by combining WSI and flow cytometry information~\cite{hashimoto2024multimodal}. These studies
show the value of multi-expert structures for modeling tissue heterogeneity, multi-task correlations, or multi-modal evidence. However, representative pathology MoE methods mainly operate after patch features have been extracted~\cite{wu2025pamoe,li2025m4,hashimoto2024multimodal}. In contrast, this work introduces MoE into high-level transformer FFN transformation for encoder-level representation formation, making it complementary to MIL-level MoE methods.
\section{Method}
\label{sec:method}

\subsection{Overview}
\label{sec:overview}
\begin{figure*}[t]
    \centering
    \includegraphics[width=\textwidth]{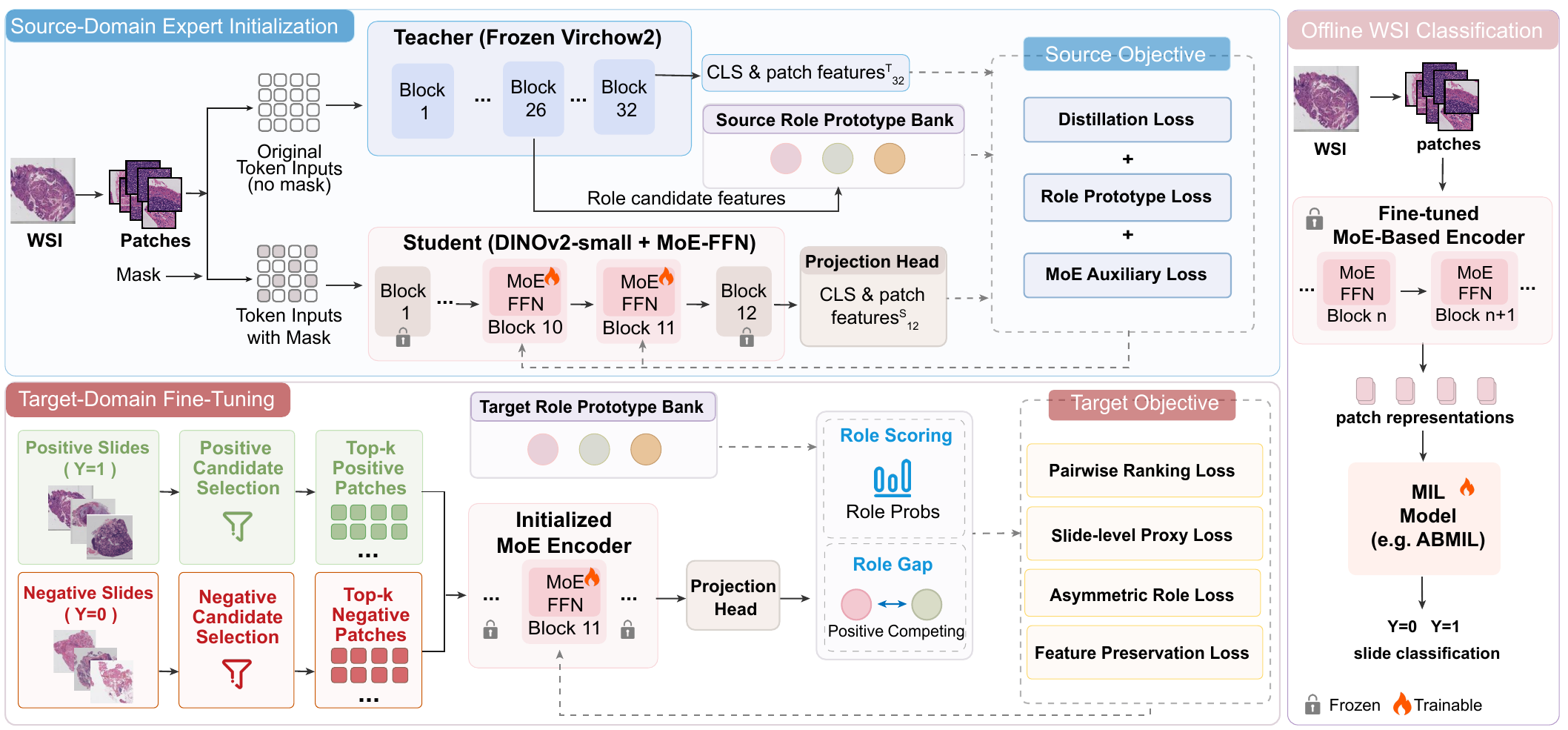}
    \caption{
    Overview of the proposed role-guided MoE framework.
    (i) Source-domain expert initialization learns pathology-aware representations and initial expert specialization using a frozen Virchow2 teacher, source role prototypes, and MoE regularization.
    (ii) Target-domain fine-tuning refines the MoE-related parameters by scoring selected positive and hard-negative patches against target role prototypes.
    (iii) Offline WSI classification fixes the fine-tuned MoE-based encoder for patch representation extraction and MIL-based slide classification.
    }
    \label{fig:training_mechanism}
\end{figure*}
To transfer pathology-aware knowledge to a lightweight MoE-based encoder and establish expert-specific transformations for downstream adaptation, we design a role-guided MoE framework, as illustrated in Fig.~\ref{fig:training_mechanism}. The framework consists of two stages of encoder-level representation learning, namely source-domain expert initialization and target-domain fine-tuning, followed by offline WSI classification.
The source stage initializes the MoE parameters using source-domain data covering multiple cancer types, enabling the model to capture general pathological patterns and establish initial expert differentiation. The target stage fine-tunes these parameters on each downstream dataset to learn task-specific patch representations.
After fine-tuning, the fixed MoE-based encoder extracts patch features, and the extracted features are aggregated by a standard MIL model for slide-level classification.

\subsubsection{Problem Formulation}
\label{sec:problem_formulation}
Let $\mathcal{D}$ denote the data used for encoder representation learning, and let $F_{\theta,\phi}$ denote the MoE-based encoder, with frozen backbone parameters $\theta$ and trainable MoE-related parameters $\phi$. Formally, the encoder representation learning problem aims to preserve pathological semantic priors while improving discrimination for the target WSI classification task. The optimal parameters are learned through the overall representation learning objective $\mathcal{L}_{\mathrm{rep}}$:
\begin{equation}
\phi^{*}
=
\min_{\phi}
\mathcal{L}_{\mathrm{rep}}
\left(
F_{\theta,\phi};\mathcal{D}
\right).
\end{equation}
Given the $i$-th WSI $S_i$, it is divided into a patch bag $X_i=\{x_{i,j}\}_{j=1}^{M_i}$ containing $M_i$ non-overlapping patches. After representation learning, each patch is encoded as
\begin{equation}
z_{i,j}
=
F_{\theta,\phi^{*}}(x_{i,j}),
\qquad j=1,\ldots,M_i.
\end{equation}
After representation learning, the encoder is fixed, and WSI classification is optimized over the MIL model $G_{\eta}$:
\begin{equation}
\min_{\eta}
\sum_{i=1}^{N}
\mathcal{L}_{\mathrm{cls}}
\left(
G_{\eta}
\left(
\{z_{i,j}\}_{j=1}^{M_i}
\right),
y_i
\right).
\end{equation}

\subsection{Source-Domain Expert Initialization} \label{sec:source_initialization} 
Source-domain expert initialization aims to build a pathology-aware expert structure before target-specific fine-tuning. By transferring general pathological priors into the MoE-FFN blocks and encouraging differentiated expert transformations, this stage provides a stable initialization for subsequent adaptation to data-limited target WSI tasks.

\subsubsection{MoE-FFN Encoder Architecture} 
\label{sec:moe_encoder}
\paragraph{MoE-FFN Integration and Expert Composition}
Standard FFNs use a single shared transformation for all tokens, which limits their ability to model heterogeneous tissue patterns. To overcome this limitation, we introduce MoE-FFNs that provide expert-specific transformations with conditional token routing. To exploit this design while preserving the pretrained representation capacity, we replace the original FFNs in selected high-level transformer blocks $\mathcal{L}_{\mathrm{moe}}$ with MoE-FFNs and keep the remaining backbone parameters frozen. We focus on high-level blocks because their representations are more likely to encode higher-order pathological morphology and tissue architecture, whereas earlier blocks mainly capture generic visual cues such as color, edges, and local textures. Accordingly, only the MoE-related components in $\mathcal{L}_{\mathrm{moe}}$, including expert FFNs, routing and normalization parameters, are optimized, focusing trainable capacity on high-level semantic transformation.

Each MoE-FFN contains one always-active shared expert for common feature transformation and $K$ routed experts that provide specialized transformations for heterogeneous tissue patterns according to token-expert compatibility.
For token $h_i^{l}$ in the $l$-th MoE layer, the output is formulated as
\begin{equation}
\label{eq:moe_output}
\operatorname{MoE}^{l}(h_i^{l})
=
\alpha E_s^{l}(h_i^{l})
+
\sum_{k=1}^{K}
g_{ik}^{l}E_k^{l}(h_i^{l}),
\end{equation}
where $E_s^{l}$ and $E_k^{l}$ denote the shared expert and the $k$-th routed expert, respectively; $\alpha$ controls the contribution of the shared expert, and $g_{ik}^{l}$ denotes the sparse routing weight defined in the following \textit{Routing Strategy} paragraph.

\paragraph{Routing Strategy}
Since a patch token may encode multiple tissue patterns, we adopt a top-any routing strategy~\cite{guo2025dynamic} that dynamically activates at most $K$ experts according to token-level complexity. 
For token $h_i^l$ in the $l$-th MoE layer, each routed expert
$E_k^l$ is represented by a learnable routing vector $v_k^l$. Let $s_{ik}^l$ denote their cosine similarity score with a learnable scale. Each expert additionally has a learnable threshold $\tau_k^l$, and the adjusted routing score is defined as
\begin{equation}
q_{ik}^{l}=s_{ik}^{l}-\tau_k^l.
\end{equation}

Let $k_{i,a}^{l}$ denote the expert with the $a$-th highest routing score for token $i$. We first keep the top-$A_{\max}$ candidate experts and then activate only those with sufficiently competitive responses:
\begin{equation}
\mathcal{K}_i^l
=
\{k_{i,a}^{l}\}_{a=1}^{A_{\max}},
\end{equation}
\begin{equation}
\mathcal{S}_i^l
=
\{k_{i,1}^{l}\}
\cup
\left\{
k_{i,a}^{l}\in\mathcal{K}_i^l
\,\middle|\,
a>1,\ 
q_{i,k_{i,a}^{l}}^{l}
\ge
\rho q_{i,k_{i,1}^{l}}^{l}
\right\}.
\end{equation}
where $\rho$ specifies the minimum relative score required for activating an additional expert. Consequently, $\lvert\mathcal{S}_i^l\rvert \in \{1,\ldots,A_{\max}\}$, bounding computation while allowing adaptive routing.
The routing weights are normalized as:
\begin{equation}
\{g_{ik}^{l}\}_{k\in\mathcal{S}_i^l}
=
\operatorname{softmax}
\left(
\{q_{ik}^{l}\}_{k\in\mathcal{S}_i^l}
\right),
\quad
g_{ik}^{l}=0\ \text{for}\ k\notin\mathcal{S}_i^l.
\end{equation}
Eq.~\eqref{eq:moe_output} uses the shared expert and selected routed experts.

\paragraph{Plug-and-Play MoE Module and Backbone Compatibility}
The two selected high-level transformer blocks equipped with MoE-FFNs constitute a plug-and-play MoE module. Once encoder-level optimization is completed, the resulting module can replace the corresponding high-level blocks in another transformer-based backbone. For backbones with different hidden dimensions, lightweight input and output projections align the backbone features with the module space. This modular design enables cross-backbone transfer without redesign.

\subsubsection{Source Role Prototype Construction} \label{sec:source_prototypes}
To guide initial expert specialization, we construct a source role prototype bank in the teacher feature space, where each prototype serves as a weak pathological anchor for a recurring tissue pattern in the source data. During source-domain training, tokens routed to each role-guided expert are encouraged to align with the corresponding prototype, promoting expert differentiation without assigning fixed tissue identities.

Since tissue-level annotations are typically unavailable in WSI datasets, role-specific candidate patches cannot be directly identified from annotated regions. We therefore employ CONCH~\cite{lu2024conch}, a pathology vision-language model with strong image-text alignment, to assign patch-level labels and relevance scores. For the source data, we define a prompt bank $\mathcal{T}^{\mathrm{src}}$ containing $R$ tissue roles.
Given a source patch $x_n^{\mathrm{src}}$, CONCH produces a predicted role label $\hat{r}_n^{\mathrm{src}}$ and the corresponding confidence score
$s_n^{\mathrm{src}}$:
\begin{equation}
\left(
\hat{r}_n^{\mathrm{src}},
s_n^{\mathrm{src}}
\right)
=
\operatorname{CONCH}
\left(
x_n^{\mathrm{src}};
\mathcal{T}^{\mathrm{src}}
\right).
\end{equation}

For each role, patches are ranked according to their confidence scores $s_n^{\mathrm{src}}$, and only the top-ranked fraction is retained to reduce the risk of introducing inaccurate pathological priors into prototype construction. The selected patches are further balanced across source organs to prevent a single organ from dominating the resulting prototype. The candidate set for role $r$ is denoted by $\mathcal{H}_r^{\mathrm{src}}$.

For each candidate patch
$x_n^{\mathrm{src}}\in\mathcal{H}_r^{\mathrm{src}}$, let $t_{n,j}^{\mathrm{T}}$ denote its $j$-th patch-token feature
extracted from layer $\ell_{\mathrm{T}}$ of the frozen teacher encoder. The patch-level feature $f_n^{\mathrm{T}}$ is obtained by mean-pooling its $J$ patch-token features:
$f_n^{\mathrm{T}}
=\operatorname{MeanPool}
(\{t_{n,j}^{\mathrm{T}}\}_{j=1}^{J})$.

To reduce the influence of noisy or morphologically atypical
candidates, the L2-normalized teacher features $\{f_n^{\mathrm{T}}\}$ of each role are clustered using K-means.
Let $\mathcal{I}_r^{*}$ denote the indices of the largest cluster containing at least $m_{\min}$ candidates. The source prototype of role $r$ is then computed as
\begin{equation}
p_r^{\mathrm{src}}
=
\frac{1}{|\mathcal{I}_r^{*}|}
\sum_{n\in\mathcal{I}_r^{*}}
f_n^{\mathrm{T}}.
\end{equation}
When no valid cluster is available, all candidate features of that role are averaged. Then, the source role prototype bank is defined as
\begin{equation}
P^{\mathrm{src}}
=
\left\{
p_r^{\mathrm{src}}
\right\}_{r=1}^{R}.
\end{equation}

\subsubsection{Source-Domain Training Objectives} 
\label{sec:source_objectives} 
To impose source-domain supervision on the MoE-based encoder, we use
DINOv2-small~\cite{oquabdinov2} as a compact student encoder and a frozen Virchow2 encoder~\cite{vorontsov2024virchow} as the pathology-aware teacher. The teacher also defines the feature space of the source role prototype bank $P^{\mathrm{src}}$. To resolve the teacher-student dimension mismatch, a projection head maps student features to the teacher feature space before supervision. The objective combines teacher-student distillation, role prototype supervision, and MoE auxiliary losses.

\paragraph{Teacher-Student Distillation Loss}
\label{sec:distillation_loss}

To transfer pathology-aware global (i.e., CLS tokens) and patch-level (i.e., patch tokens) representations, we align the final-layer token features of the student encoder and the frozen teacher, as features from this layer directly form the encoder output, while the student features also capture the representation changes produced by the optimized MoE blocks. Let $z_{\mathrm{cls}}^{\mathrm{S}}$ and $\{z_i^{\mathrm{S}}\}_{i=1}^{N}$ denote the student CLS and patch tokens, and let$z_{\mathrm{cls}}^{\mathrm{T}}$ and $\{z_i^{\mathrm{T}}\}_{i=1}^{N}$ denote the corresponding teacher tokens. The projection head $W_p$ maps both $z_{\mathrm{cls}}^{\mathrm{T}}$ and $\{z_i^{\mathrm{T}}\}_{i=1}^{N}$ into the teacher feature space. The projected student tokens are denoted by
$\tilde{z}^{\mathrm{S}}$. 

Then, we align the student and teacher CLS tokens using the cosine distance loss
$\mathcal{L}_{\mathrm{cls}}
=1-\cos(\tilde{z}_{\mathrm{cls}}^{\mathrm{S}},
z_{\mathrm{cls}}^{\mathrm{T}})$
to preserve global semantic consistency.

To encourage the student to infer pathology-aware tissue representations from surrounding context rather than relying only on directly visible patch content, we apply block masking to the student tokens while the teacher processes the complete image. 
Let $\mathcal{M}$ and $\mathcal{U}$ denote the masked and unmasked patch-token index sets, respectively, with alignment losses
\begin{align}
\mathcal{L}_{\mathrm{mask}}
&=
\frac{1}{|\mathcal{M}|}
\sum_{i\in\mathcal{M}}
\operatorname{SmoothL1}
\left(
\tilde{z}_{i}^{\mathrm{S}},
z_{i}^{\mathrm{T}}
\right),
\\
\mathcal{L}_{\mathrm{unmask}}
&=
\frac{1}{|\mathcal{U}|}
\sum_{i\in\mathcal{U}}
\operatorname{SmoothL1}
\left(
\tilde{z}_{i}^{\mathrm{S}},
z_{i}^{\mathrm{T}}
\right).
\end{align}

Beyond individual token alignment, we randomly sample token pairs $\Omega$. For each pair in $\Omega$, we encourage the difference between the two student token features to approximate that between the corresponding teacher token features:
\begin{equation}
\mathcal{L}_{\mathrm{rel}}
=
\frac{1}{|\Omega|}
\sum_{(i,j)\in\Omega}
\left\|
\left(
\tilde{z}_{i}^{\mathrm{S}}
-
\tilde{z}_{j}^{\mathrm{S}}
\right)
-
\left(
z_{i}^{\mathrm{T}}
-
z_{j}^{\mathrm{T}}
\right)
\right\|_2^2.
\end{equation}

The complete distillation objective is
\begin{equation}
\mathcal{L}_{\mathrm{distill}}
=
\lambda_{\mathrm{cls}}\mathcal{L}_{\mathrm{cls}}
+
\lambda_{\mathrm{mask}}\mathcal{L}_{\mathrm{mask}}
+
\lambda_{\mathrm{unmask}}\mathcal{L}_{\mathrm{unmask}}
+
\lambda_{\mathrm{rel}}\mathcal{L}_{\mathrm{rel}},
\end{equation}

\paragraph{Role Prototype Loss}
\label{sec:source_role_loss}
To promote pathology-relevant expert differentiation, we associate $R$ of the $K$ routed experts with the $R$ source role prototypes and retain one unconstrained free expert, such that $K=R+1$. The free expert captures patterns beyond the predefined roles.

\begin{figure}
    \centering
    \includegraphics[width=1\linewidth]{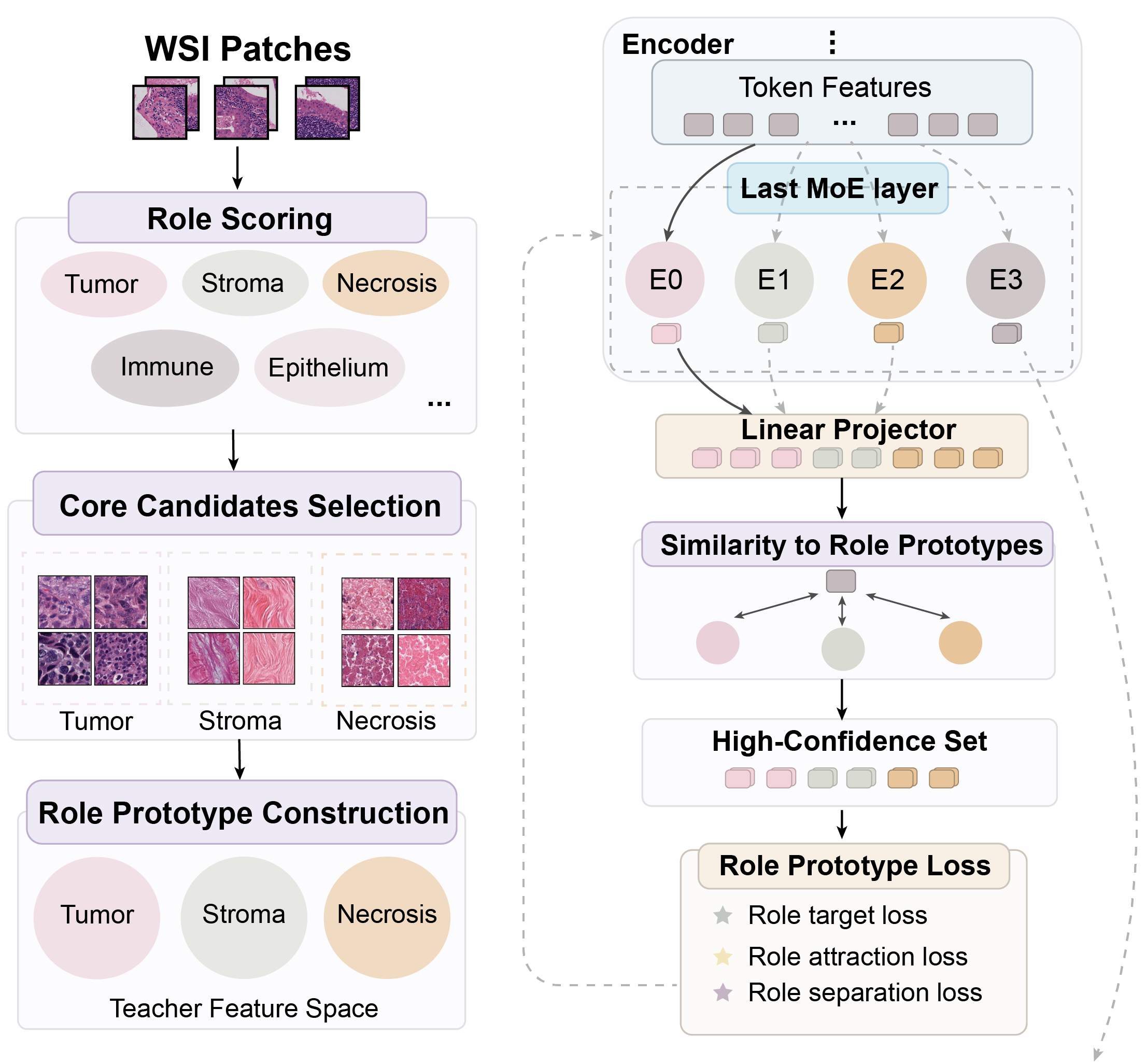}
    \caption{
    Role prototype-guided supervision.
    Left: candidate selection and prototype construction.
    Right: token alignment with role prototypes.
    }
    \label{fig:role_proto_supervision}
\end{figure}

As illustrated in Fig.~\ref{fig:role_proto_supervision}, role
supervision is applied to a token from the last block in
$\mathcal{L}_{\mathrm{moe}}$ only when a role-guided expert receives the highest routing weight for that token. After projection into the teacher feature space, the similarity score $s_{i,r}$ between token $i$ and source role prototype $r$ is computed as their cosine similarity. The role prototype probabilities are computed as $\mathbf{a}_i=\operatorname{softmax}(\mathbf{s}_i/\tau)$, where $\tau$ is the temperature.

To reduce supervision noise, we retain only tokens with confident prototype assignments, determined by the highest prototype probability and its margin over the second highest. Let $\mathcal{V}_r^{\mathrm{src}}$ denote the high-confidence tokens dominated by the role-guided expert associated with role $r$.

The role target loss aligns each selected token with its corresponding role prototype:
\begin{equation}
\mathcal{L}_{\mathrm{target}}^{(r)}
=
\mathbb{E}_{i\in\mathcal{V}_r^{\mathrm{src}}}
\left[
-\log a_{i,r}
\right].
\end{equation}

The role attraction loss pulls each selected token toward its corresponding
role prototype:
\begin{equation}
\mathcal{L}_{\mathrm{attr}}^{(r)}
=
\mathbb{E}_{i\in\mathcal{V}_r^{\mathrm{src}}}
\left[
1-s_{i,r}
\right].
\end{equation}

Let $s_{i,\neg r}=\max_{q\neq r}s_{i,q}$ denote the highest similarity between token $i$ and all role prototypes except its corresponding prototype $r$. The role separation loss requires the similarity to prototype $r$ to exceed this value
by a margin $m$:
\begin{equation}
\mathcal{L}_{\mathrm{sep}}^{(r)}
=
\frac{1}{|\mathcal{V}_r^{\mathrm{src}}|}
\sum_{i\in\mathcal{V}_r^{\mathrm{src}}}
\max\left(
0,\,
m+s_{i,\neg r}-s_{i,r}
\right).
\end{equation}

Let $w_r = |\mathcal{V}_r^{\mathrm{src}}| / \sum_{q=1}^{R} |\mathcal{V}_q^{\mathrm{src}}|$ denote the proportion of high-confidence tokens associated with role $r$.
The complete source role prototype loss is
\begin{equation}
\mathcal{L}_{\mathrm{role}}^{\mathrm{src}}
=
\sum_{r=1}^{R}
w_r
\bigl(
\lambda_{t,r}\mathcal{L}_{\mathrm{target}}^{(r)}
+
\lambda_{a,r}\mathcal{L}_{\mathrm{attr}}^{(r)}
+
\lambda_{s,r}\mathcal{L}_{\mathrm{sep}}^{(r)}
\bigr).
\end{equation}

\paragraph{MoE Auxiliary Loss}
\label{sec:moe_auxiliary_loss}

To stabilize expert routing, we use four auxiliary losses. Following the principle in sparse MoE models~\cite{shazeer2017outrageously,fedus2022switch}, the load balancing loss $\mathcal{L}_{\mathrm{bal}}$ promotes balanced token assignment and routing weights across routed experts. Following DynMoE~\cite{guo2025dynamic}, the diversity loss $\mathcal{L}_{\mathrm{div}}$ separates the normalized routing vectors. We further introduce two losses to promote reliable and sparse expert routing. The coverage loss ensures that each token receives a strong response from at least one routed expert. The sparsity loss controls routing complexity by constraining the average number of activated experts per token:
\begin{align}
\mathcal{L}_{\mathrm{cov}}
&=
\frac{1}{|\mathcal{T}|}
\sum_{i\in\mathcal{T}}
\max(0,c-\max_k s_{i,k}),
\\
\mathcal{L}_{\mathrm{spr}}
&=
\left(
\mathbb{E}_{i\in\mathcal{T}}[a_i]
-a^*
\right)^2.
\end{align}
where $c$ is the minimum routing score threshold, and $a^{*}$ is the target average number of activated experts per token.

The four auxiliary terms are computed for each MoE layer and then averaged across all selected MoE layers. The resulting MoE auxiliary loss is
\begin{equation}
\mathcal{L}_{\mathrm{aux}}
=
\lambda_{\mathrm{div}}\mathcal{L}_{\mathrm{div}}
+
\lambda_{\mathrm{bal}}\mathcal{L}_{\mathrm{bal}}
+
\lambda_{\mathrm{cov}}\mathcal{L}_{\mathrm{cov}}
+
\lambda_{\mathrm{spr}}\mathcal{L}_{\mathrm{spr}}.
\end{equation}

\paragraph{Overall Source Objective} 
\label{sec:overall_source_objective}
The source-domain training jointly transfers pathology-aware representations from the teacher, promotes role-guided expert differentiation, and stabilizes expert routing. The overall source objective is
\begin{equation}
\mathcal{L}_{src}
=
\lambda_{distill}\mathcal{L}_{distill}
+
\lambda_{role}^{src}\mathcal{L}_{role}^{src}
+
\lambda_{aux}\mathcal{L}_{aux}.
\end{equation}
where the three coefficients balance the distillation, role prototype, and MoE auxiliary losses.

\subsection{Target-Domain Fine-Tuning}
\label{sec:target_finetuning}

The source-initialized experts may not fully match the tissue composition and discriminative cues of a specific downstream task, especially when positive patterns resemble confusing negative regions. Starting from the source-initialized encoder, we construct a task-specific target role prototype bank $P^{\mathrm{tgt}}$ for each downstream dataset. We then further optimize the MoE-related parameters to enhance target-specific discrimination while preserving the source-initialized representations.

\subsubsection{Target Role Prototype Construction}
\label{sec:target_prototypes}

Following the procedure described in Sec.~\ref{sec:source_prototypes}, we construct a task-specific target role prototype bank
\begin{equation}
P^{\mathrm{tgt}}
=
\left\{p_r^{\mathrm{tgt}}\right\}_{r=1}^{R}
\end{equation}
from representative patches in the corresponding target dataset.

We designate the prototype corresponding to the tissue pattern indicative of the positive class as $r^{+}$, and define $\mathcal{R}_{\mathrm{comp}}$ as the set of all remaining role prototypes. These remaining prototypes serve as competing references to the positive role.

\subsubsection{Positive and Hard-Negative Candidate Selection}
\label{sec:target_candidate_selection}

Since slide-level labels provide no patch-level localization, applying role supervision to all patches may introduce noise. We therefore mine positive and hard-negative patches for target-domain fine-tuning.

Using the same projection and similarity computation as in source-domain role supervision, we obtain the similarity $s_i^r$ and role probability for each target patch $i$ and target prototype $r$. We use the positive role gap to measure how strongly patch $i$ favors the positive role over the competing roles:
\begin{equation}
g_i
=
s_i^{r^{+}}
-
\max_{r\in\mathcal{R}_{\mathrm{comp}}} s_i^{r}.
\end{equation}

Let $\mathcal{B}^{+}$ and $\mathcal{B}^{-}$ denote the sets of positive and negative slides, respectively. For $b\in\mathcal{B}^{+}$, let $\Omega_b^{+}$ contain patches with sufficiently high positive-role probability and role gap. For $b\in\mathcal{B}^{-}$, let $\Omega_b^{-}$ contain patches with strong positive-role responses but limited neighborhood support. Let
$q_i^{\mathrm{role}}$ measure direct positive-role evidence, $q_i^{\mathrm{ctx}}$ measure neighborhood-supported positive evidence, and $q_i^{\mathrm{hard}}$ measure the likelihood that a negative patch is confused with the positive pattern. The initial candidate pools, which define the search space for target-domain fine-tuning, are constructed as
\begin{align}
\mathcal{C}_b^{+}
&=
\operatorname{TopK}_{i\in\Omega_b^{+}}
\left(q_i^{\mathrm{role}}\right)
\cup
\operatorname{TopK}_{i\in\Omega_b^{+}}
\left(q_i^{\mathrm{ctx}}\right),
\quad b\in\mathcal{B}^{+},
\\
\mathcal{C}_b^{-}
&=
\operatorname{TopK}_{i\in\Omega_b^{-}}
\left(q_i^{\mathrm{hard}}\right),
\quad b\in\mathcal{B}^{-}.
\end{align}

\subsubsection{Target-Domain Training Objectives}
\label{sec:target_objectives}

Target-domain fine-tuning combines patch-level discrimination, slide-level consistency, and representation preservation. For each slide $b$, let $\mathcal{C}_b$ denote its pre-constructed candidate pool, where $\mathcal{C}_b=\mathcal{C}_b^{+}$ for a positive slide and $\mathcal{C}_b=\mathcal{C}_b^{-}$ for a negative slide. During fine-tuning, the current encoder re-scores the patches in $\mathcal{C}_b$ and dynamically selects a subset $\mathcal{S}_b\subseteq\mathcal{C}_b$.

\paragraph{Asymmetric Role Loss}
\label{sec:asymmetric_role_loss}

The asymmetric role loss increases the positive role gap for positive candidates while suppressing it for hard-negative candidates. Let $\mathcal{S}^{+}$ and $\mathcal{S}^{-}$ denote the selected candidates from positive and negative slides, respectively. Let $m_{\mathrm{pos}}$ and $-m_{\mathrm{neg}}$ denote the desired lower and upper bounds of the positive role gap for positive and hard-negative candidates. Let $\omega_i$ denote the confidence weight derived from the role response and neighborhood context of candidate $i$. 
The loss is defined as
\begin{equation}
\begin{aligned}
\mathcal{L}_{\mathrm{asym}}
={}&
\frac{
\sum_{i\in\mathcal{S}^{+}}
\omega_i
\max\left(0,m_{\mathrm{pos}}-g_i\right)
}{
\sum_{i\in\mathcal{S}^{+}}\omega_i
}
\\
&+
\frac{
\sum_{i\in\mathcal{S}^{-}}
\omega_i
\max\left(0,m_{\mathrm{neg}}+g_i\right)
}{
\sum_{i\in\mathcal{S}^{-}}\omega_i
}.
\end{aligned}
\end{equation}

\paragraph{Pairwise Ranking Loss}
\label{sec:ranking_loss}

The ranking loss encourages positive candidates to have larger role gaps than negative candidates. Let $\bar{g}_{K}^{+}$ and $\bar{g}_{K}^{-}$ denote the mean top-$K$ role gaps of the selected positive and negative candidates, and let $m_{\mathrm{rank}}$ denote the ranking margin. The loss is defined as
\begin{equation}
\mathcal{L}_{\mathrm{rank}}
=
\max\left(
0,\,
m_{\mathrm{rank}}
-\bar{g}_{K}^{+}
+\bar{g}_{K}^{-}
\right).
\end{equation}

\paragraph{Slide-Level Proxy Loss}
\label{sec:proxy_loss}

The slide-level proxy loss aligns the strongest patch-level evidence
with the corresponding slide label. For each slide $b$, we rank the candidates in $\mathcal{S}_b$ by their role gaps and retain the top $K_b$ candidates as $\mathcal{T}_b$, where $K_b=\min(K,|\mathcal{S}_b|)$. Let $\bar{g}_b$ denote the mean role gap of the top-$K_b$ candidates in slide $b$. Let $y_b$ denote the slide label and $B$ the number of slides. The loss is defined as
\begin{equation}
\mathcal{L}_{\mathrm{proxy}}
=
\frac{1}{B}
\sum_{b=1}^{B}
\ell_{\mathrm{BCE}}
\left(
\sigma(\bar{g}_b),y_b
\right).
\end{equation}

\paragraph{Feature Preservation Loss}
\label{sec:feature_preservation_loss}

The feature preservation loss limits representation drift during
target-domain fine-tuning. For each patch in the mined candidate pool,
it minimizes the cosine distance between the representation
$h_{b,i}^{\mathrm{tgt}}$ produced by the current encoder and the
reference representation $h_{b,i}^{\mathrm{src}}$ produced by the
frozen source-initialized encoder:
\begin{equation}
\mathcal{L}_{\mathrm{pres}}
=
\mathbb{E}_{b}
\mathbb{E}_{i\in\mathcal{C}_b}
\left[
1-\cos
\left(
h_{b,i}^{\mathrm{tgt}},
h_{b,i}^{\mathrm{src}}
\right)
\right].
\end{equation}

\paragraph{Overall Target Objective}
\label{sec:overall_target_objective}
The target-domain objective jointly improves patch-level role
discrimination, preserves slide-level label consistency, and limits representation drift from the source-initialized encoder. The overall target-domain fine-tuning objective is
\begin{equation}
\mathcal{L}_{\mathrm{tgt}}
=
\lambda_{\mathrm{asym}}
\mathcal{L}_{\mathrm{asym}}
+
\lambda_{\mathrm{rank}}
\mathcal{L}_{\mathrm{rank}}
+
\lambda_{\mathrm{proxy}}
\mathcal{L}_{\mathrm{proxy}}
+
\lambda_{\mathrm{pres}}
\mathcal{L}_{\mathrm{pres}}.
\end{equation}

\section{Experiments and Results}
\label{sec:experiments}

\subsection{Experimental Setup}
\label{sec:experimental_setup}

\subsubsection{Datasets}
\label{sec:datasets}
The study uses source-domain data for expert initialization and two target-domain WSI datasets for downstream evaluation. Source-domain training includes diagnostic H\&E-stained WSIs from TCGA-BRCA, TCGA-COAD, TCGA-KIRC, and TCGA-LUAD~\cite{tcga2013pancancer}. To reduce cancer-type imbalance, we sampled equal numbers of tumor and non-tumor slides from each cohort. HISTAI-SPIDER, a public multi-organ patch dataset, was additionally included to increase the diversity of tissue patterns during source-domain training~\cite{nechaev2025spider}.

Target-domain evaluation was conducted on the public BRACS dataset and a private PAROTID dataset. For BRACS~\cite{brancati2022bracs}, we followed the official WSI-level split and used the Group\_BT versus Group\_AT binary classification task, where Group\_BT denotes benign tumors and Group\_AT denotes atypical tumors. This task involves
fine-grained breast lesions with confusing morphological patterns.

The private PAROTID dataset contains 195 H\&E-stained WSIs of parotid tumors collected from Shenzhen People's Hospital, Guangdong Province, for benign versus malignant classification. Slide-level labels were obtained from pathological diagnoses and reviewed by experienced pathologists, and all WSIs were de-identified before analysis. The dataset was divided into training, validation, and test sets at the patient level with a ratio of 0.70, 0.15, and 0.15, ensuring that all slides from the same patient remained in the same split.

\subsubsection{Evaluation Metrics}
\label{sec:evaluation_metrics}

For all datasets, we use the Area Under Curve (AUC) and macro F1-score (F1) to evaluate classification performance. All methods are evaluated using the same data split and downstream training protocol. Each experiment is repeated with five random seeds, and the mean and standard deviation are reported.

\subsubsection{Baselines and Implementation Details}
\label{sec:implementation_details}
We conduct experiments across five backbones, including UNI~\cite{chen2024uni}, UNI2-h~\cite{chen2024uni}, Virchow2~\cite{vorontsov2024virchow}, DINOv2-small~\cite{oquabdinov2}, and OpenCLIP ViT-B/16~\cite{cherti2023reproducible}, and two MIL aggregators, ABMIL~\cite{ilse2018attention} and TransMIL~\cite{shao2021transmil}. Under each backbone-MIL setting, we compare the proposed method with four baselines: (1) Frozen, which directly uses pretrained encoder for offline feature extraction without encoder updates; (2) Partial FT~\cite{lee2025benchmarking}, which fine-tunes selected high-level encoder blocks; (3) LoRA~\cite{hu2022lora}, which applies low-rank adaptation with a limited number of trainable parameters; and (4) R$^2$T~\cite{tang2024feature}, which re-embeds frozen patch features through a feature refinement module.

For the main DINOv2-small configuration, the FFNs in the 10th and 11th transformer blocks are replaced with MoE-FFNs. Each MoE-FFN contains four routed experts and one shared expert, with up to two routed experts activated for each token.
AdamW is used as the optimizer for all encoder-level training stages. Source-domain initialization is performed for 15 epochs with a weight decay of 0.05; the learning rate is initialized at $1\times10^{-4}$ and reduced to $5\times10^{-5}$ for the final 5 epochs. Target-domain fine-tuning is performed for 10 epochs with a learning rate of $2\times10^{-5}$ and a weight decay of $1\times10^{-4}$.

For downstream WSI classification, each encoder extracts offline features from 1024 patches per WSI, which are subsequently aggregated by ABMIL or TransMIL. 
The hyperparameters of Partial FT and LoRA were selected through validation experiments, and the best-performing configurations were used for comparison. R$^2$T was implemented using the recommended settings from the original paper. Under the same backbone-MIL setting, all encoder variants use identical downstream architectures and training protocols.

\subsection{Downstream WSI Classification Results}
\label{sec:classification_results}

\begin{table*}[t]
\centering
\caption{
Main comparison across backbones and MIL aggregators. AUC and F1 are reported for each dataset. Bold and underlined values denote the best and second-best results within each backbone-aggregator setting.
}

\label{tab:main_comparison}
\scriptsize
\setlength{\tabcolsep}{2.2pt}
\renewcommand{\arraystretch}{0.98}
\resizebox{\textwidth}{!}{
\begin{tabular}{llcccccccc}
\toprule
\multirow{3}{*}{Backbone}
& \multirow{3}{*}{Method}
& \multicolumn{4}{c}{ABMIL}
& \multicolumn{4}{c}{TransMIL} \\
\cmidrule(lr){3-6} \cmidrule(lr){7-10}
& & \multicolumn{2}{c}{PAROTID} & \multicolumn{2}{c}{BRACS}
  & \multicolumn{2}{c}{PAROTID} & \multicolumn{2}{c}{BRACS} \\
\cmidrule(lr){3-4} \cmidrule(lr){5-6} \cmidrule(lr){7-8} \cmidrule(lr){9-10}
& & AUC & F1 & AUC & F1 & AUC & F1 & AUC & F1 \\
\midrule

\multirow{5}{*}{UNI}
& Frozen
& $0.898 \pm 0.031$ & $0.725 \pm 0.097$
& $0.649 \pm 0.068$ & $0.546 \pm 0.143$
& $0.957 \pm 0.019$ & $0.875 \pm 0.030$
& $0.681 \pm 0.054$ & $\underline{0.627 \pm 0.091}$ \\

& Partial FT
& $0.903 \pm 0.059$ & $0.819 \pm 0.038$
& $0.514 \pm 0.008$ & $0.540 \pm 0.036$
& $0.967 \pm 0.018$ & $0.853 \pm 0.085$
& $0.456 \pm 0.057$ & $0.380 \pm 0.129$ \\

& LoRA
& $0.925 \pm 0.029$ & $\underline{0.822 \pm 0.043}$
& $0.651 \pm 0.068$ & $0.551 \pm 0.142$
& $\underline{0.970 \pm 0.018}$ & $0.845 \pm 0.063$
& $0.625 \pm 0.066$ & $0.585 \pm 0.049$ \\

& R$^2$T
& $\mathbf{0.962 \pm 0.005}$ & $0.787 \pm 0.019$
& $\underline{0.730 \pm 0.008}$ & $\underline{0.674 \pm 0.014}$
& $0.961 \pm 0.016$ & $\underline{0.902 \pm 0.015}$
& $\underline{0.700 \pm 0.027}$ & $0.608 \pm 0.024$ \\

& Ours
& $\underline{0.954 \pm 0.026}$ & $\mathbf{0.860 \pm 0.034}$
& $\mathbf{0.765 \pm 0.026}$ & $\mathbf{0.693 \pm 0.045}$
& $\mathbf{0.982 \pm 0.010}$ & $\mathbf{0.933 \pm 0.015}$
& $\mathbf{0.720 \pm 0.030}$ & $\mathbf{0.636 \pm 0.066}$ \\

\midrule

\multirow{5}{*}{UNI2-h}
& Frozen
& $0.964 \pm 0.015$ & $0.871 \pm 0.078$
& $0.751 \pm 0.018$ & $\underline{0.612 \pm 0.031}$
& $0.970 \pm 0.020$ & $0.840 \pm 0.028$
& $0.639 \pm 0.041$ & $\underline{0.526 \pm 0.051}$ \\

& Partial FT
& $0.978 \pm 0.012$ & $\underline{0.939 \pm 0.037}$
& $0.751 \pm 0.019$ & $0.605 \pm 0.034$
& $0.971 \pm 0.005$ & $0.842 \pm 0.076$
& $0.573 \pm 0.073$ & $0.345 \pm 0.250$ \\

& LoRA
& $\underline{0.979 \pm 0.012}$ & $0.933 \pm 0.032$
& $0.752 \pm 0.019$ & $\underline{0.612 \pm 0.031}$
& $\underline{0.980 \pm 0.018}$ & $\underline{0.896 \pm 0.052}$
& $0.582 \pm 0.056$ & $0.365 \pm 0.170$ \\

& R$^2$T
& $0.961 \pm 0.001$ & $0.913 \pm 0.013$
& $\underline{0.755 \pm 0.019}$ & $0.548 \pm 0.033$
& $0.963 \pm 0.002$ & $0.788 \pm 0.000$
& $\mathbf{0.716 \pm 0.080}$ & $0.485 \pm 0.184$ \\

& Ours
& $\mathbf{0.985 \pm 0.008}$ & $\mathbf{0.940 \pm 0.014}$
& $\mathbf{0.776 \pm 0.015}$ & $\mathbf{0.722 \pm 0.026}$
& $\mathbf{0.983 \pm 0.011}$ & $\mathbf{0.946 \pm 0.018}$
& $\underline{0.711 \pm 0.033}$ & $\mathbf{0.595 \pm 0.056}$ \\

\midrule

\multirow{5}{*}{Virchow2}
& Frozen
& $\underline{0.936 \pm 0.043}$ & $0.760 \pm 0.062$
& $0.785 \pm 0.033$ & $0.669 \pm 0.059$
& $0.930 \pm 0.030$ & $\underline{0.839 \pm 0.114}$
& $0.708 \pm 0.077$ & $0.604 \pm 0.063$ \\

& Partial FT
& $0.667 \pm 0.005$ & $0.608 \pm 0.082$
& $\mathbf{0.786 \pm 0.030}$ & $0.671 \pm 0.051$
& $0.609 \pm 0.041$ & $0.490 \pm 0.183$
& $0.692 \pm 0.058$ & $0.591 \pm 0.069$ \\

& LoRA
& $0.914 \pm 0.081$ & $\underline{0.768 \pm 0.086}$
& $\underline{0.786 \pm 0.032}$ & $\underline{0.677 \pm 0.049}$
& $\underline{0.963 \pm 0.019}$ & $0.836 \pm 0.128$
& $\mathbf{0.741 \pm 0.033}$ & $\underline{0.622 \pm 0.046}$ \\

& R$^2$T
& $0.932 \pm 0.011$ & $0.737 \pm 0.007$
& $0.745 \pm 0.021$ & $0.581 \pm 0.037$
& $0.923 \pm 0.014$ & $0.774 \pm 0.054$
& $0.728 \pm 0.029$ & $0.614 \pm 0.102$ \\

& Ours
& $\mathbf{0.970 \pm 0.014}$ & $\mathbf{0.853 \pm 0.018}$
& $0.782 \pm 0.027$ & $\mathbf{0.683 \pm 0.015}$
& $\mathbf{0.969 \pm 0.011}$ & $\mathbf{0.884 \pm 0.026}$
& $\underline{0.737 \pm 0.024}$ & $\mathbf{0.639 \pm 0.037}$ \\

\midrule

\multirow{5}{*}{DINOv2-s}
& Frozen
& $0.833 \pm 0.054$ & $0.665 \pm 0.100$
& $0.458 \pm 0.043$ & $0.323 \pm 0.104$
& $0.782 \pm 0.061$ & $0.533 \pm 0.244$
& $0.459 \pm 0.074$ & $\underline{0.426 \pm 0.112}$ \\

& Partial FT
& $0.832 \pm 0.049$ & $\mathbf{0.774 \pm 0.038}$
& $0.447 \pm 0.018$ & $0.322 \pm 0.182$
& $0.826 \pm 0.032$ & $0.681 \pm 0.055$
& $0.496 \pm 0.092$ & $0.419 \pm 0.157$ \\

& LoRA
& $0.835 \pm 0.037$ & $0.683 \pm 0.126$
& $\underline{0.480 \pm 0.099}$ & $0.000 \pm 0.000$
& $\underline{0.829 \pm 0.055}$ & $0.670 \pm 0.104$
& $\underline{0.528 \pm 0.070}$ & $0.000 \pm 0.000$ \\

& R$^2$T
& $\underline{0.878 \pm 0.015}$ & $0.762 \pm 0.037$
& $0.474 \pm 0.035$ & $\underline{0.405 \pm 0.025}$
& $\mathbf{0.853 \pm 0.022}$ & $\underline{0.701 \pm 0.053}$
& $0.458 \pm 0.020$ & $0.390 \pm 0.125$ \\

& Ours
& $\mathbf{0.890 \pm 0.011}$ & $\underline{0.763 \pm 0.014}$
& $\mathbf{0.530 \pm 0.015}$ & $\mathbf{0.484 \pm 0.097}$
& $0.802 \pm 0.041$ & $\mathbf{0.712 \pm 0.045}$
& $\mathbf{0.607 \pm 0.026}$ & $\mathbf{0.459 \pm 0.042}$ \\

\midrule

\multirow{5}{*}{OpenCLIP}
& Frozen
& $0.899 \pm 0.050$ & $0.810 \pm 0.074$
& $\underline{0.434 \pm 0.023}$ & $0.404 \pm 0.056$
& $0.895 \pm 0.009$ & $0.760 \pm 0.107$
& $0.413 \pm 0.026$ & $0.330 \pm 0.199$ \\

& Partial FT
& $0.907 \pm 0.049$ & $\underline{0.822 \pm 0.068}$
& $0.426 \pm 0.018$ & $0.404 \pm 0.064$
& $0.897 \pm 0.011$ & $0.763 \pm 0.110$
& $\underline{0.452 \pm 0.074}$ & $0.328 \pm 0.097$ \\

& LoRA
& $\underline{0.909 \pm 0.046}$ & $\underline{0.822 \pm 0.068}$
& $0.432 \pm 0.023$ & $0.347 \pm 0.113$
& $0.899 \pm 0.014$ & $0.758 \pm 0.105$
& $0.435 \pm 0.035$ & $\underline{0.413 \pm 0.146}$ \\

& R$^2$T
& $0.909 \pm 0.008$ & $0.779 \pm 0.071$
& $0.421 \pm 0.032$ & $\underline{0.413 \pm 0.052}$
& $\underline{0.916 \pm 0.004}$ & $\underline{0.802 \pm 0.033}$
& $0.409 \pm 0.033$ & $0.377 \pm 0.085$ \\

& Ours
& $\mathbf{0.911 \pm 0.019}$ & $\mathbf{0.833 \pm 0.013}$
& $\mathbf{0.443 \pm 0.011}$ & $\mathbf{0.448 \pm 0.053}$
& $\mathbf{0.928 \pm 0.016}$ & $\mathbf{0.856 \pm 0.026}$
& $\mathbf{0.486 \pm 0.035}$ & $\mathbf{0.478 \pm 0.039}$ \\

\bottomrule
\end{tabular}
}
\end{table*}

Table~\ref{tab:main_comparison} reports the AUC and F1 scores on PAROTID and BRACS across all backbone-MIL settings. Overall, our method achieves the best F1 score in 19 of the 20 settings and ranks first or second in AUC in 19 settings, showing consistent improvements across backbones and MIL configurations.

On PAROTID, where only limited target-domain training data are available, our method obtains the best F1 score in nine of the ten settings, indicating stable improvement in a low-data setting. The only exception is DINOv2-small with ABMIL, where Partial FT achieves an F1 score of 0.774 compared with 0.763 for our method. Nevertheless, our method achieves the highest AUC of 0.890 in this setting, improving upon Partial FT by 0.058. In comparison, the relative performance of the baseline strategies varies more noticeably across backbones. On BRACS, our method achieves the best F1 score in all ten settings. The largest improvement is observed for UNI2-h with ABMIL, where the F1 score increases from 0.612 for the best competing method to 0.722 for our method. These results suggest that the proposed method produces more discriminative patch representations for the fine-grained classification of benign and atypical lesions.

Finally, the similar performance gains observed with both ABMIL and TransMIL suggest that the improvements mainly arise from enhanced patch representations rather than dependence on a particular MIL model.

\subsection{Ablation Studies}
\label{sec:component_ablation}
To analyze the contribution of each component, we conduct ablation studies on UNI, as shown in Table~\ref{tab:ablation}.

\textbf{MoE architecture.}
Adding a randomly initialized MoE improves the F1 score of Frozen UNI from 0.725 to 0.791 on PAROTID and from 0.546 to 0.585 on BRACS. This result shows that the additional expert branches provide useful task-specific transformation capacity. However, random MoE remains below the full model by 0.069 and 0.108 in F1 on the two datasets, respectively, indicating that the MoE architecture alone does not account for the overall improvement.

\textbf{Distillation.}
The full model achieves the highest AUC and F1 score on both datasets. Among the ablated variants, removing distillation causes the largest performance reduction. Compared with the full model, the AUC decreases by 0.023 on PAROTID and 0.056 on BRACS, while the F1 score decreases by 0.060 and 0.054, respectively. This result supports the importance of transferring pathology-aware representation priors from the teacher encoder.

\textbf{Role prototype guidance.}
Removing role prototype guidance decreases AUC and F1 on both datasets, with a more evident reduction in F1. This degradation indicates that role guidance provides information beyond the MoE structure and distillation by encouraging distinct expert transformations associated with task-relevant tissue patterns.

\textbf{Target-domain fine-tuning.}
 On PAROTID, removing target-domain fine-tuning leaves the AUC unchanged at 0.954 but reduces F1 from 0.860 to 0.800. The unchanged AUC suggests that source-domain initialization already provides a useful overall ranking of positive and negative slides, whereas target-domain fine-tuning improves the final classification decisions under the target data distribution. Its smaller effect on BRACS suggests that the source-initialized representations transfer more directly to this dataset, while further target-specific optimization is more beneficial for PAROTID.

\begin{table*}[t]
\centering
\caption{
Ablation study of the proposed role-guided MoE encoder adaptation framework using UNI as the backbone.
}
\label{tab:ablation}
\resizebox{\textwidth}{!}{
\begin{tabular}{lcccccccc}
\toprule
\multirow{2}{*}{Method}
& \multirow{2}{*}{MoE}
& \multirow{2}{*}{Distill.}
& \multirow{2}{*}{Role proto.}
& \multirow{2}{*}{Target FT}
& \multicolumn{2}{c}{PAROTID}
& \multicolumn{2}{c}{BRACS} \\
\cmidrule(lr){6-7}
\cmidrule(lr){8-9}
& & & & & AUC & F1 & AUC & F1 \\
\midrule
Frozen UNI
&  &  &  & 
& $0.898 \pm 0.031$ & $0.725 \pm 0.097$
& $0.649 \pm 0.068$ & $0.546 \pm 0.143$ \\

UNI + random MoE
& \checkmark &  &  & 
& $0.928 \pm 0.028$ & $0.791 \pm 0.036$
& $0.687 \pm 0.111$ & $0.585 \pm 0.134$ \\

Ours w/o distillation
& \checkmark &  & \checkmark & \checkmark
& $0.931 \pm 0.041$ & $\underline{0.800 \pm 0.048}$
& $0.709 \pm 0.083$ & $0.639 \pm 0.104$ \\

Ours w/o role prototype
& \checkmark & \checkmark &  & \checkmark
& $0.949 \pm 0.029$ & $0.800 \pm 0.087$
& $0.753 \pm 0.035$ & $0.685 \pm 0.035$ \\

Ours w/o target fine-tuning
& \checkmark & \checkmark & \checkmark & 
& $\underline{0.954 \pm 0.027}$ & $0.800 \pm 0.087$
& $\underline{0.759 \pm 0.038}$ & $\underline{0.690 \pm 0.036}$ \\

Ours (full)
& \checkmark & \checkmark & \checkmark & \checkmark
& $\mathbf{0.954 \pm 0.026}$ & $\mathbf{0.860 \pm 0.034}$
& $\mathbf{0.765 \pm 0.026}$ & $\mathbf{0.693 \pm 0.045}$ \\
\bottomrule
\end{tabular}
}
\end{table*}

\subsection{Hyperparameter Analysis}
\label{sec:hyperparameter_analysis}

\begin{figure}
    \centering
    \includegraphics[width=1\linewidth]{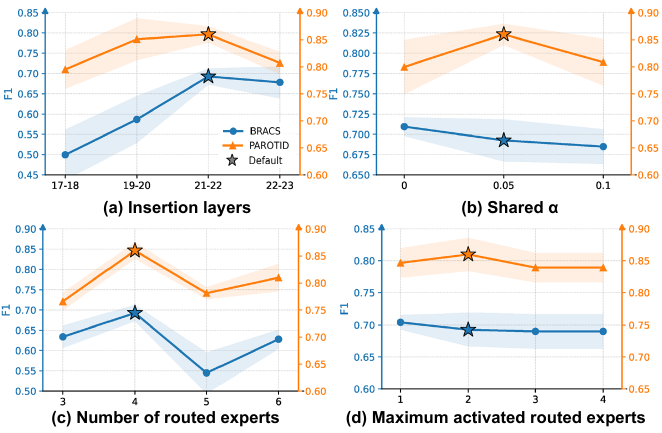}
    \caption{F1-based hyperparameter analysis of the proposed method.}
    \label{fig:hyperparameter_analysis}
\end{figure}

We conduct hyperparameter analysis using UNI with ABMIL by varying one factor at a time while keeping the remaining settings fixed. The F1 results are shown in Fig.~\ref{fig:hyperparameter_analysis}.

\textbf{MoE-FFN insertion layers.}
On UNI, performance generally improves as the MoE-FFNs are moved from earlier to higher transformer blocks, reaching the highest F1 at layers 21-22 on both datasets, followed by a decline at layers 22-23. Based on this trend, we adopt a relative insertion rule across backbones, placing the MoE-FFNs in selected high-level blocks while leaving the final block unchanged.

\textbf{Shared-expert weight.}
BRACS performs best at $\alpha=0$, whereas PAROTID reaches its highest F1 at $\alpha=0.05$. Since $\alpha=0.05$ substantially improves PAROTID while maintaining competitive performance on BRACS, it is selected as the default trade-off across the two datasets.

\textbf{Number of routed experts.}
When varying the number of routed experts, one routed expert is kept free, while the remaining experts are guided by role prototypes. F1 improves when the number of routed experts increases from three to four but decreases with five or six experts. We therefore use four routed experts as the default setting.

\textbf{Maximum activated routed experts.}
The performance is relatively stable when the maximum number of activated routed experts is set to one or two. BRACS slightly favors one expert, whereas PAROTID achieves its best performance with two. Increasing the upper bound to three or four provides no further gains. We therefore use two as the default to retain adaptive expert collaboration while maintaining stable performance across both datasets.

\subsection{Representation and Downstream Evidence Analysis}
\label{sec:representation_analysis}

\begin{figure}
    \centering
    \includegraphics[width=1\linewidth]{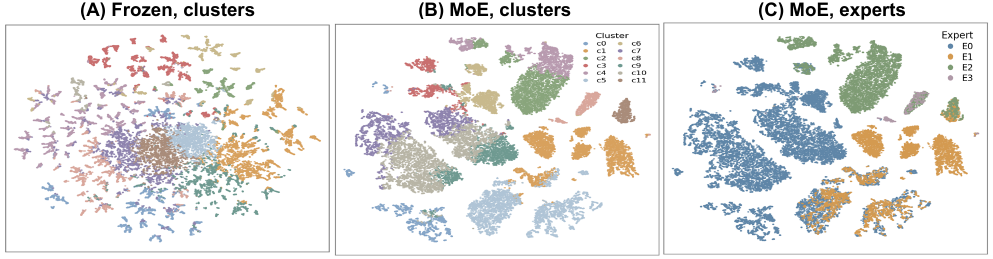}
    \caption{
    t-SNE visualization of frozen and MoE-based DINOv2-small features, colored by clusters in (A,B) and expert assignments in (C).
    }
    \label{fig:tsne_last_moe}
\end{figure}

To examine how the proposed method changes patch representations, we visualize features from the last MoE layer (i.e., the 11th block of DINOv2-small) using t-SNE. Frozen DINOv2-small features are extracted from the same block for comparison. As shown in Fig.~\ref{fig:tsne_last_moe}, the frozen features are broadly dispersed with substantial overlap between clusters. In contrast, the features produced by the MoE-based encoder form more compact groups with clearer separation, indicating a more structured representation space.
Panels (B) and (C) show the same MoE feature embedding colored by token cluster and routed expert, respectively. Several clusters are dominated by particular experts, while some regions contain contributions from multiple experts. This pattern suggests that the routing mechanism organizes token features into expert-associated subspaces without forcing a strict one-to-one correspondence between experts and tissue patterns. Together, these observations provide qualitative evidence that the MoE encoder reorganizes high-level token representations and promotes complementary expert specialization.

\begin{figure}
    \centering
    \includegraphics[width=\linewidth]{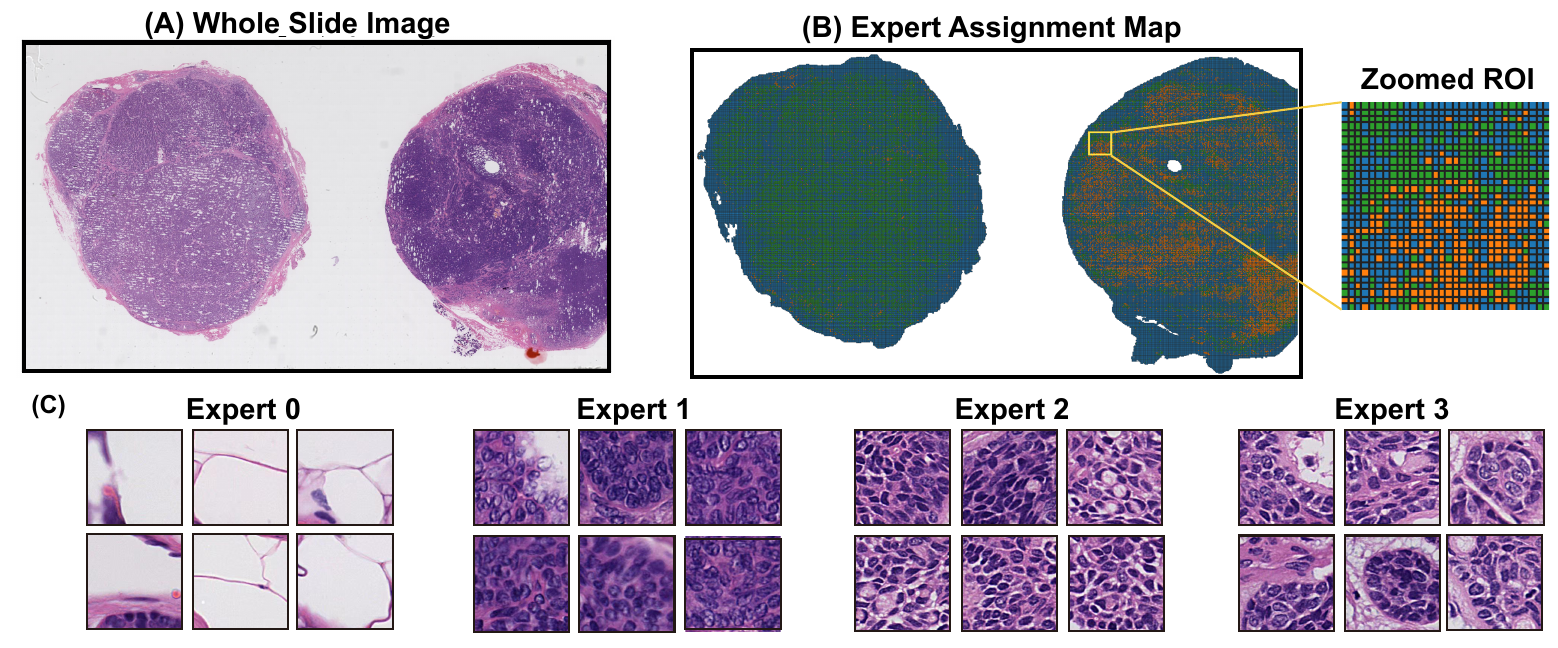}
    \caption{
    Morphological interpretation of expert preferences based on patch-level expert composition.
    (A) WSI F24-0328A01\_H01 from the PAROTID dataset.
    (B) Expert assignment map and zoomed ROI.
    (C) Representative high-response patches for each expert.
    }
    \label{fig:expert_interpretation}
\end{figure}

Fig.~\ref{fig:expert_interpretation} provides a morphological interpretation of the expert preferences. The expert assignment map shows spatially heterogeneous routing across the WSI. Among the representative high-response patches, Expert 0 is mainly associated with non-tumor and background regions, including fibroadipose tissue, whereas Experts 1-3 respond predominantly to tumor-rich regions with different local architectures, such as solid and gland-like patterns. These observations show that different experts emphasize different morphological patterns, supporting complementary specialization within the MoE encoder.

\begin{figure}
    \centering
    \includegraphics[width=\linewidth]{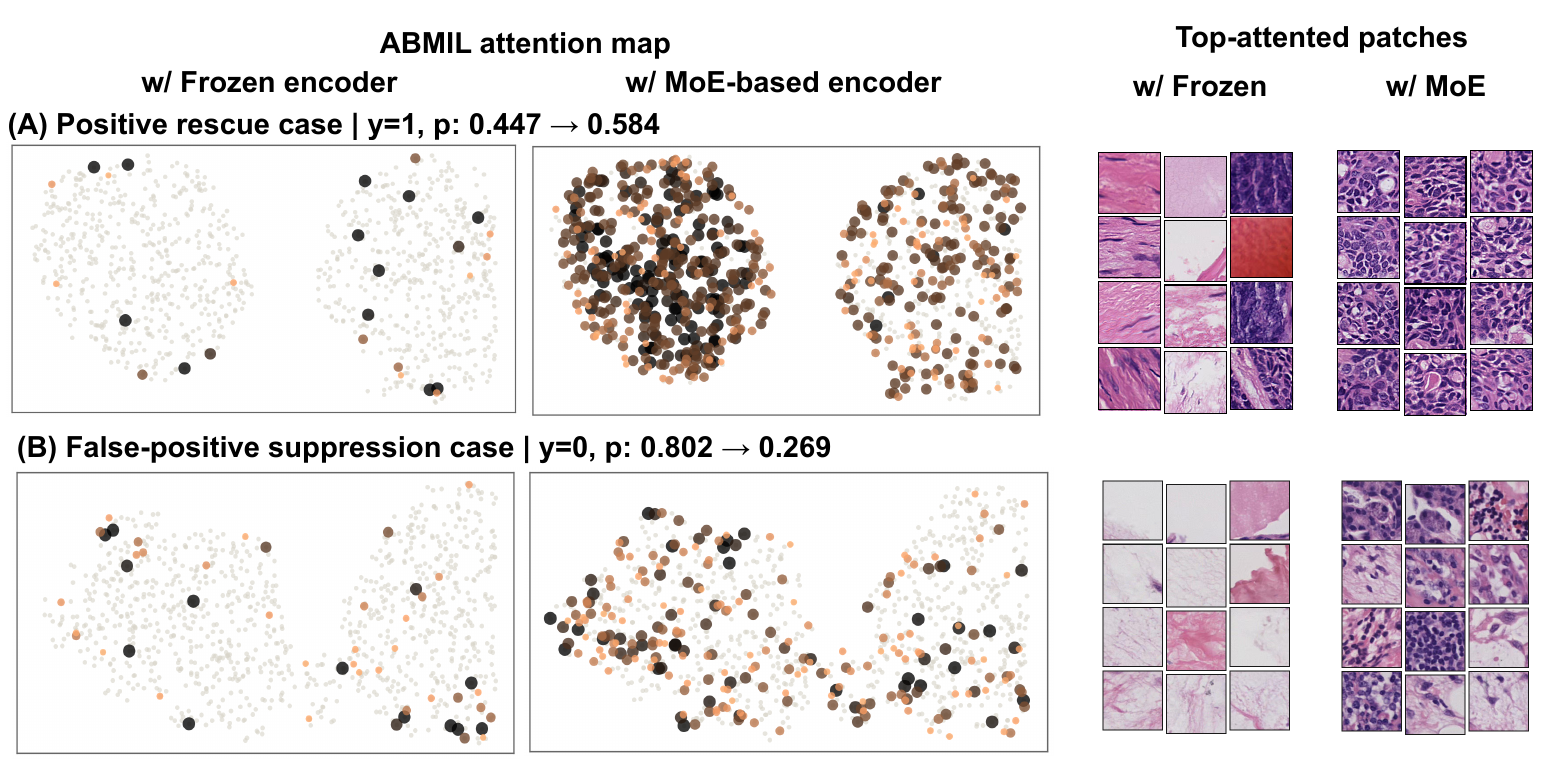}
    \caption{
    Downstream evidence shift after MoE optimization. ABMIL attention maps and top-attended patches show changes in slide-level evidence selection.
    }
    \label{fig:downstream_evidence_shift}
\end{figure}

We further examine whether the differences in encoder representations affect the evidence selected by the downstream MIL model. As shown in Fig.~\ref{fig:downstream_evidence_shift}, the MoE-based features alter both the ABMIL attention distribution and the resulting high-attention patches. In the positive rescue case, attention is shifted toward tumor-rich regions, increasing the predicted positive probability from 0.447 to 0.584. In the false-positive suppression case, the selected evidence becomes more consistent with benign tissue patterns, reducing the positive probability from 0.802 to 0.269. These examples show that changes in patch representations propagate to the MIL attention mechanism and influence which tissue regions are used for slide-level prediction.

\section{Discussion and Conclusion}
\label{sec:discussion_conclusion}

In MIL-based WSI classification, downstream performance is often constrained by frozen patch representations, while direct encoder fine-tuning may overfit on limited WSI data, and standard shared transformations may be insufficient for heterogeneous tissue patterns. To address these limitations, we propose a pathology role-guided MoE-FFN framework for efficient encoder-level representation learning. The framework introduces transformation diversity into selected high-level transformer blocks, transfers pathology-aware priors through teacher-student distillation, and uses role prototypes to promote initial expert specialization. Target-domain fine-tuning further refines the initialized experts with asymmetric prototype-guided optimization, improving discrimination between task-relevant positive patterns and confusable hard negatives. As a plug-and-play module, the optimized MoE-FFN blocks can be transferred across transformer-based backbones.
Experiments on the public BRACS dataset and the private PAROTID dataset demonstrate consistent improvements across five backbones and two MIL aggregators. Representation analyses further show a more structured token feature space, distinct expert preferences for tissue morphology, and shifted MIL evidence selection. These results suggest that the proposed framework improves target-aware patch representation quality and benefits slide-level WSI classification.

\bibliographystyle{IEEEtran}

\section*{References}

\IEEEtriggercmd{\vspace*{-15pt}}
\IEEEtriggeratref{1}

\bibliography{references}

@IEEEtranBSTCTL{BSTcontrol,
  CTLuse_forced_etal       = "yes",
  CTLmax_names_forced_etal = "6",
  CTLnames_show_etal       = "1"
}

@article{campanella2019clinical,
    title={Clinical-grade computational pathology using weakly supervised deep learning on whole slide images},
  author={Campanella, Gabriele and Hanna, Matthew G and Geneslaw, Luke and Miraflor, Allen and Werneck Krauss Silva, Vitor and Busam, Klaus J and Brogi, Edi and Reuter, Victor E and Klimstra, David S and Fuchs, Thomas J},
  journal={Nature medicine},
  volume={25},
  number={8},
  pages={1301--1309},
  year={2019},
  publisher={Nature Publishing Group US New York}
}

@inproceedings{ilse2018attention,
  title={Attention-based deep multiple instance learning},
  author={Ilse, Maximilian and Tomczak, Jakub and Welling, Max},
  booktitle={International conference on machine learning},
  pages={2127--2136},
  year={2018},
  organization={PMLR}
}

@article{lu2021dataefficient,
  title={Data-efficient and weakly supervised computational pathology on whole-slide images},
  author={Lu, Ming Y and Williamson, Drew FK and Chen, Tiffany Y and Chen, Richard J and Barbieri, Matteo and Mahmood, Faisal},
  journal={Nature biomedical engineering},
  volume={5},
  number={6},
  pages={555--570},
  year={2021},
  publisher={Nature Publishing Group UK London}
}

@article{shao2021transmil,
  title={Transmil: Transformer based correlated multiple instance learning for whole slide image classification},
  author={Shao, Zhuchen and Bian, Hao and Chen, Yang and Wang, Yifeng and Zhang, Jian and Ji, Xiangyang and others},
  journal={Advances in neural information processing systems},
  volume={34},
  pages={2136--2147},
  year={2021}
}

@article{chen2024uni,
  title={Towards a general-purpose foundation model for computational pathology},
  author={Chen, Richard J and Ding, Tong and Lu, Ming Y and Williamson, Drew FK and Jaume, Guillaume and Song, Andrew H and Chen, Bowen and Zhang, Andrew and Shao, Daniel and Shaban, Muhammad and others},
  journal={Nature medicine},
  volume={30},
  number={3},
  pages={850--862},
  year={2024},
  publisher={Nature Publishing Group US New York}
}

@article{vorontsov2024virchow,
  title={A foundation model for clinical-grade computational pathology and rare cancers detection},
  author={Vorontsov, Eugene and Bozkurt, Alican and Casson, Adam and Shaikovski, George and Zelechowski, Michal and Severson, Kristen and Zimmermann, Eric and Hall, James and Tenenholtz, Neil and Fusi, Nicolo and others},
  journal={Nature medicine},
  volume={30},
  number={10},
  pages={2924--2935},
  year={2024},
  publisher={Nature Publishing Group US New York}
}

@article{lu2024conch,
  title={A visual-language foundation model for computational pathology},
  author={Lu, Ming Y and Chen, Bowen and Williamson, Drew FK and Chen, Richard J and Liang, Ivy and Ding, Tong and Jaume, Guillaume and Odintsov, Igor and Le, Long Phi and Gerber, Georg and others},
  journal={Nature medicine},
  volume={30},
  number={3},
  pages={863--874},
  year={2024},
  publisher={Nature Publishing Group US New York}
}

@inproceedings{tang2024feature,
  title={Feature re-embedding: Towards foundation model-level performance in computational pathology},
  author={Tang, Wenhao and Zhou, Fengtao and Huang, Sheng and Zhu, Xiang and Zhang, Yi and Liu, Bo},
  booktitle={Proceedings of the IEEE/CVF conference on computer vision and pattern recognition},
  pages={11343--11352},
  year={2024}
}

@article{oquabdinov2,
  title={DINOv2: Learning Robust Visual Features without Supervision},
  author={Oquab, Maxime and Darcet, Timoth{\'e}e and Moutakanni, Th{\'e}o and Vo, Huy and Szafraniec, Marc and Khalidov, Vasil and Fernandez, Pierre and Haziza, Daniel and Massa, Francisco and El-Nouby, Alaaeldin and others},
  journal={Transactions on Machine Learning Research Journal},
  year={2024}
}

@inproceedings{touvron2021training,
  title={Training data-efficient image transformers \& distillation through attention},
  author={Touvron, Hugo and Cord, Matthieu and Douze, Matthijs and Massa, Francisco and Sablayrolles, Alexandre and J{\'e}gou, Herv{\'e}},
  booktitle={International conference on machine learning},
  pages={10347--10357},
  year={2021},
  organization={PMLR}
}

@article{snell2017prototypical,
  title={Prototypical networks for few-shot learning},
  author={Snell, Jake and Swersky, Kevin and Zemel, Richard},
  journal={Advances in neural information processing systems},
  volume={30},
  year={2017}
}

@article{wang2022ctranspath,
  title={Transformer-based unsupervised contrastive learning for histopathological image classification},
  author={Wang, Xiyue and Yang, Sen and Zhang, Jun and Wang, Minghui and Zhang, Jing and Yang, Wei and Huang, Junzhou and Han, Xiao},
  journal={Medical image analysis},
  volume={81},
  pages={102559},
  year={2022},
  publisher={Elsevier}
}

@article{xu2024gigapath,
  title={A whole-slide foundation model for digital pathology from real-world data},
  author={Xu, Hanwen and Usuyama, Naoto and Bagga, Jaspreet and Zhang, Sheng and Rao, Rajesh and Naumann, Tristan and Wong, Cliff and Gero, Zelalem and Gonz{\'a}lez, Javier and Gu, Yu and others},
  journal={Nature},
  volume={630},
  number={8015},
  pages={181--188},
  year={2024},
  publisher={Nature Publishing Group UK London}
}

@inproceedings{houlsby2019parameter,
  title={Parameter-efficient transfer learning for NLP},
  author={Houlsby, Neil and Giurgiu, Andrei and Jastrzebski, Stanislaw and Morrone, Bruna and De Laroussilhe, Quentin and Gesmundo, Andrea and Attariyan, Mona and Gelly, Sylvain},
  booktitle={International conference on machine learning},
  pages={2790--2799},
  year={2019},
  organization={PMLR}
}

@inproceedings{hu2022lora,
  author       = {Edward J. Hu and
                  Yelong Shen and
                  Phillip Wallis and
                  Zeyuan Allen-Zhu and
                  Yuanzhi Li and
                  Shean Wang and
                  Lu Wang and
                  Weizhu Chen},
  title        = {LoRA: Low-Rank Adaptation of Large Language Models},
  booktitle    = {The Tenth International Conference on Learning Representations, {ICLR}
                  2022, Virtual Event, April 25-29, 2022},
  year         = {2022}
}

@article{liu2026hiadapter,
    author = {Liu, Qingyang and Xie, Peng and Dai, Zhehao and Bai, Xiangzhi},
    year = {2026},
    month = {05},
    pages = {1-1},
    title = {HiAdapter: Histopathology-induced Adapter for Pathology Foundation Models},
    journal = {IEEE Transactions on Medical Imaging},
    doi = {10.1109/TMI.2026.3694387}
}

@article{fedus2022switch,
  title={Switch transformers: Scaling to trillion parameter models with simple and efficient sparsity},
  author={Fedus, William and Zoph, Barret and Shazeer, Noam},
  journal={Journal of Machine Learning Research},
  volume={23},
  number={120},
  pages={1--39},
  year={2022}
}

@article{riquelme2021scaling,
  title={Scaling vision with sparse mixture of experts},
  author={Riquelme, Carlos and Puigcerver, Joan and Mustafa, Basil and Neumann, Maxim and Jenatton, Rodolphe and Susano Pinto, Andr{\'e} and Keysers, Daniel and Houlsby, Neil},
  journal={Advances in Neural Information Processing Systems},
  volume={34},
  pages={8583--8595},
  year={2021}
}

@inproceedings{wu2025pamoe,
  title={Learning heterogeneous tissues with mixture of experts for gigapixel whole slide images},
  author={Wu, Junxian and Chen, Minheng and Ke, Xinyi and Xun, Tianwang and Jiang, Xiaoming and Zhou, Hongyu and Shao, Lizhi and Kong, Youyong},
  booktitle={Proceedings of the Computer Vision and Pattern Recognition Conference},
  pages={5144--5153},
  year={2025}
}

@article{li2025m4,
  title={M4: Multi-proxy multi-gate mixture of experts network for multiple instance learning in histopathology image analysis},
  author={Li, Junyu and Zhang, Ye and Shu, Wen and Feng, Xiaobing and Wang, Yingchun and Yan, Pengju and Li, Xiaolin and Sha, Chulin and He, Min},
  journal={Medical Image Analysis},
  volume={103},
  pages={103561},
  year={2025},
  publisher={Elsevier}
}

@article{hashimoto2024multimodal,
  title={Multimodal gated mixture of experts using whole slide image and flow cytometry for multiple instance learning classification of lymphoma},
  author={Hashimoto, Noriaki and Hanada, Hiroyuki and Miyoshi, Hiroaki and Nagaishi, Miharu and Sato, Kensaku and Hontani, Hidekata and Ohshima, Koichi and Takeuchi, Ichiro},
  journal={Journal of Pathology Informatics},
  volume={15},
  pages={100359},
  year={2024},
  publisher={Elsevier}
}

@article{ding2025titan,
  title={A multimodal whole-slide foundation model for pathology},
  author={Ding, Tong and Wagner, Sophia J and Song, Andrew H and Chen, Richard J and Lu, Ming Y and Zhang, Andrew and Vaidya, Anurag J and Jaume, Guillaume and Shaban, Muhammad and Kim, Ahrong and others},
  journal={Nature medicine},
  pages={1--13},
  year={2025},
  publisher={Nature Publishing Group US New York}
}

@article{tcga2013pancancer,
  title={The cancer genome atlas pan-cancer analysis project},
  author={Weinstein, John N and Collisson, Eric A and Mills, Gordon B and Shaw, Kenna R and Ozenberger, Brad A and Ellrott, Kyle and Shmulevich, Ilya and Sander, Chris and Stuart, Joshua M},
  journal={Nature genetics},
  volume={45},
  number={10},
  pages={1113--1120},
  year={2013},
  publisher={Nature Publishing Group}
}

@article{brancati2022bracs,
  title={Bracs: A dataset for breast carcinoma subtyping in h\&e histology images},
  author={Brancati, Nadia and Anniciello, Anna Maria and Pati, Pushpak and Riccio, Daniel and Scognamiglio, Giosu{\`e} and Jaume, Guillaume and De Pietro, Giuseppe and Di Bonito, Maurizio and Foncubierta, Antonio and Botti, Gerardo and others},
  journal={Database},
  volume={2022},
  pages={baac093},
  year={2022},
  publisher={Oxford University Press UK}
}

@article{nechaev2025spider,
  title={SPIDER: a comprehensive multi-organ supervised pathology dataset and baseline models},
  author={Nechaev, Dmitry and Pchelnikov, Alexey and Ivanova, Ekaterina},
  journal={arXiv preprint arXiv:2503.02876},
  year={2025}
}

@article{jacobs1991adaptive,
  title={Adaptive mixtures of local experts},
  author={Jacobs, Robert A and Jordan, Michael I and Nowlan, Steven J and Hinton, Geoffrey E},
  journal={Neural computation},
  volume={3},
  number={1},
  pages={79--87},
  year={1991},
  publisher={MIT Press}
}

@article{huang2024free,
  title={Free lunch in pathology foundation model: Task-specific model adaptation with concept-guided feature enhancement},
  author={Huang, Yanyan and Zhao, Weiqin and Chen, Yihang and Fu, Yu and Yu, Lequan},
  journal={Advances in Neural Information Processing Systems},
  volume={37},
  pages={79963--79995},
  year={2024}
}

@inproceedings{lu2024pathotune,
  title={Pathotune: Adapting visual foundation model to pathological specialists},
  author={Lu, Jiaxuan and Yan, Fang and Zhang, Xiaofan and Gao, Yue and Zhang, Shaoting},
  booktitle={International Conference on Medical Image Computing and Computer-Assisted Intervention},
  pages={395--406},
  year={2024},
  organization={Springer}
}

@inproceedings{kang2023benchmarking,
  title={Benchmarking self-supervised learning on diverse pathology datasets},
  author={Kang, Mingu and Song, Heon and Park, Seonwook and Yoo, Donggeun and Pereira, S{\'e}rgio},
  booktitle={Proceedings of the IEEE/CVF Conference on Computer Vision and Pattern Recognition},
  pages={3344--3354},
  year={2023}
}

@article{howard2021impact,
  title={The impact of site-specific digital histology signatures on deep learning model accuracy and bias},
  author={Howard, Frederick M and Dolezal, James and Kochanny, Sara and Schulte, Jefree and Chen, Heather and Heij, Lara and Huo, Dezheng and Nanda, Rita and Olopade, Olufunmilayo I and Kather, Jakob N and others},
  journal={Nature communications},
  volume={12},
  number={1},
  pages={4423},
  year={2021},
  publisher={Nature Publishing Group UK London}
}

@article{lee2025benchmarking,
  title={Benchmarking pathology foundation models: Adaptation strategies and scenarios},
  author={Lee, Jaeung and Lim, Jeewoo and Byeon, Keunho and Kwak, Jin Tae},
  journal={Computers in Biology and Medicine},
  volume={190},
  pages={110031},
  year={2025},
  publisher={Elsevier}
}

@inproceedings{dai2024deepseekmoe,
  title={Deepseekmoe: Towards ultimate expert specialization in mixture-of-experts language models},
  author={Dai, Damai and Deng, Chengqi and Zhao, Chenggang and Xu, RX and Gao, Huazuo and Chen, Deli and Li, Jiashi and Zeng, Wangding and Yu, Xingkai and Wu, Yu and others},
  booktitle={Proceedings of the 62nd Annual Meeting of the Association for Computational Linguistics (Volume 1: Long Papers)},
  pages={1280--1297},
  year={2024}
}

@article{chi2022representation,
  title={On the representation collapse of sparse mixture of experts},
  author={Chi, Zewen and Dong, Li and Huang, Shaohan and Dai, Damai and Ma, Shuming and Patra, Barun and Singhal, Saksham and Bajaj, Payal and Song, Xia and Mao, Xian-Ling and others},
  journal={Advances in Neural Information Processing Systems},
  volume={35},
  pages={34600--34613},
  year={2022}
}

@inproceedings{guo2025dynamic,
  title={Dynamic mixture of experts: An auto-tuning approach for efficient transformer models},
  author={Guo, Yongxin and Cheng, Zhenglin and Tang, Xiaoying and Tu, Zhaopeng and Lin, Tao},
  booktitle={International Conference on Learning Representations},
  volume={2025},
  pages={79643--79672},
  year={2025}
}

@inproceedings{shazeer2017outrageously,
  title={Outrageously Large Neural Networks: The Sparsely-Gated Mixture-of-Experts Layer},
  author={Shazeer, Noam and Mirhoseini, Azalia and Maziarz, Krzysztof and Davis, Andy and Le, Quoc and Hinton, Geoffrey and Dean, Jeff},
  booktitle={International Conference on Learning Representations},
  year={2017}
}

@inproceedings{cherti2023reproducible,
  title={Reproducible scaling laws for contrastive language-image learning},
  author={Cherti, Mehdi and Beaumont, Romain and Wightman, Ross and Wortsman, Mitchell and Ilharco, Gabriel and Gordon, Cade and Schuhmann, Christoph and Schmidt, Ludwig and Jitsev, Jenia},
  booktitle={Proceedings of the IEEE/CVF Conference on Computer Vision and Pattern Recognition},
  pages={2818--2829},
  year={2023}
}

\end{document}